%% file: asplos_CR.tex
\documentclass[sigconf,screen]{acmart}
\AtBeginDocument{%
  \providecommand\BibTeX{{%
    \normalfont B\kern-0.5em{\scshape i\kern-0.25em b}\kern-0.8em\TeX}}}

\usepackage{amsmath}
\usepackage{graphicx}
\usepackage{caption}
\usepackage{subcaption}
\usepackage[toc,page]{appendix}
\usepackage{xcolor}

\usepackage{listings}

\definecolor{codegreen}{rgb}{0,0.6,0}
\definecolor{codegray}{rgb}{0.5,0.5,0.5}
\definecolor{codepurple}{rgb}{0.58,0,0.82}
\definecolor{backcolour}{rgb}{0.95,0.95,0.92}
\lstdefinestyle{mystyle}{
    backgroundcolor=\color{backcolour},   
    commentstyle=\color{codegreen},
    keywordstyle=\color{magenta},
    numberstyle=\tiny\color{codegray},
    stringstyle=\color{codepurple},
    basicstyle=\ttfamily\footnotesize,
    breakatwhitespace=false,         
    breaklines=true,                 
    captionpos=b,                    
    keepspaces=true,                 
    numbers=left,                    
    numbersep=5pt,                  
    showspaces=false,                
    showstringspaces=false,
    showtabs=false,                  
    tabsize=2
}
\DeclareMathOperator*{\argmin}{\arg\!\min}
\DeclareMathOperator*{\minimize}{min}
\newcommand{\fapprox}{$f^{*}$}
\newcommand{\capprox}{$c^{*}$}

\setcopyright{acmcopyright}
\acmPrice{15.00}
\acmDOI{10.1145/3445814.3446762}
\acmYear{2021}
\copyrightyear{2021}
\acmSubmissionID{asplos21main-p1450-p}
\acmISBN{978-1-4503-8317-2/21/04}
\acmConference[ASPLOS '21]{Proceedings of the 26th ACM International Conference on Architectural Support for Programming Languages and Operating Systems}{April 19--23, 2021}{Virtual, USA}
\acmBooktitle{Proceedings of the 26th ACM International Conference on Architectural Support for Programming Languages and Operating Systems (ASPLOS '21), April 19--23, 2021, Virtual, USA}

\begin{document}
    
\title{Mind Mappings: Enabling Efficient Algorithm-Accelerator Mapping Space Search}

\author{Kartik Hegde}
\affiliation{%
  \institution{University of Illinois at Urbana-Champaign, USA}
}
\email{kvhegde2@illinois.edu}

\author{Po-An Tsai}
\affiliation{%
  \institution{NVIDIA, USA}
    }
\email{poant@nvidia.com}

\author{Sitao Huang}
\affiliation{%
  \institution{University of Illinois at Urbana-Champaign, USA}
}
\email{shuang91@illinois.edu}

\author{Vikas Chandra}
\affiliation{%
  \institution{Facebook, USA}
    }
\email{vchandra@fb.com}

\author{Angshuman Parashar}
\affiliation{%
  \institution{NVIDIA, USA}
    }
\email{aparashar@nvidia.com}

\author{Christopher W. Fletcher}
\affiliation{%
  \institution{University of Illinois at Urbana-Champaign, USA}
}
\email{cwfletch@illinois.edu}

\sloppy 
\begin{abstract}
    Modern day computing increasingly relies on specialization to satiate growing performance and efficiency requirements.
    A core challenge in designing such specialized hardware architectures is how to perform \emph{mapping space search}, i.e., search for an optimal mapping from algorithm to hardware.
    Prior work shows that choosing an inefficient mapping can lead to multiplicative-factor efficiency overheads.
    Additionally, the search space is not only large but also non-convex and non-smooth, precluding advanced search techniques.
    As a result, previous works are forced to implement mapping space search using expert choices or sub-optimal search heuristics. 
    
    This work proposes \emph{Mind Mappings}, a novel gradient-based search method for algorithm-accelerator mapping space search.
    The key idea is to derive a smooth, differentiable approximation to the otherwise non-smooth, non-convex search space. 
    With a smooth, differentiable approximation, we can leverage efficient \emph{gradient-based} search algorithms to find high-quality mappings.
    We extensively compare Mind Mappings to black-box optimization schemes used in prior work.
    When tasked to find mappings for two important workloads (CNN and MTTKRP), the proposed search finds mappings that achieve an average $1.40\times$, $1.76\times$, and $1.29\times$~(when run for a fixed number of steps) and $3.16\times$, $4.19\times$, and $2.90\times$~(when run for a fixed amount of time) better energy-delay product (EDP) relative to Simulated Annealing, Genetic Algorithms and Reinforcement Learning, respectively. Meanwhile, Mind Mappings
    returns mappings with only $5.32\times$ higher EDP than a possibly unachievable theoretical lower-bound, indicating proximity to the global optima.
\end{abstract}



\maketitle




\input{ASPLOS CR/intro}
\input{ASPLOS CR/mapspace.tex}

\input{ASPLOS CR/method.tex}

\input{ASPLOS CR/evaluation.tex}

\input{ASPLOS CR/background.tex}
\input{ASPLOS CR/conclusion.tex}
\begin{acks}
We thanks the anonymous reviewers and our shepherd Daniel Jimenez for their valuable feedback.
This work was funded in part by NSF under grant 1942888 and by DARPA SDH contract DARPA SDH \#HR0011-18-3-0007.  Kartik Hegde was funded in part by a Facebook Ph.D. fellowship.
\end{acks}

\appendix

\input{ASPLOS CR/appendix.tex}

\bibliographystyle{ACM-Reference-Format}
\input{ASPLOS CR/asplos_CR.bbl}


\end{document}

%% file: ASPLOS CR/intro.tex
\section{Introduction}
\label{sec:intro}

The compound effect of the slowing of Moore's law coupled with a growing demand for efficient compute has ushered in an era of specialized hardware architectures.
Due to their inherent performance, energy, and area characteristics, these accelerators are driving innovation in diverse areas such as machine learning~\cite{eyeriss, ucnn, cnvlutin, eie, scnn, tpu}, medicine~\cite{medical1, medical2, medical3}, cryptography~\cite{crypt1, crypt2}, etc.
They are seeing a wide variety of deployments ranging from cloud to edge---forcing designers to make complex design decisions to achieve their efficiency objectives.

\begin{figure}[t]
\begin{centering}
  \includegraphics[width=0.85\columnwidth]{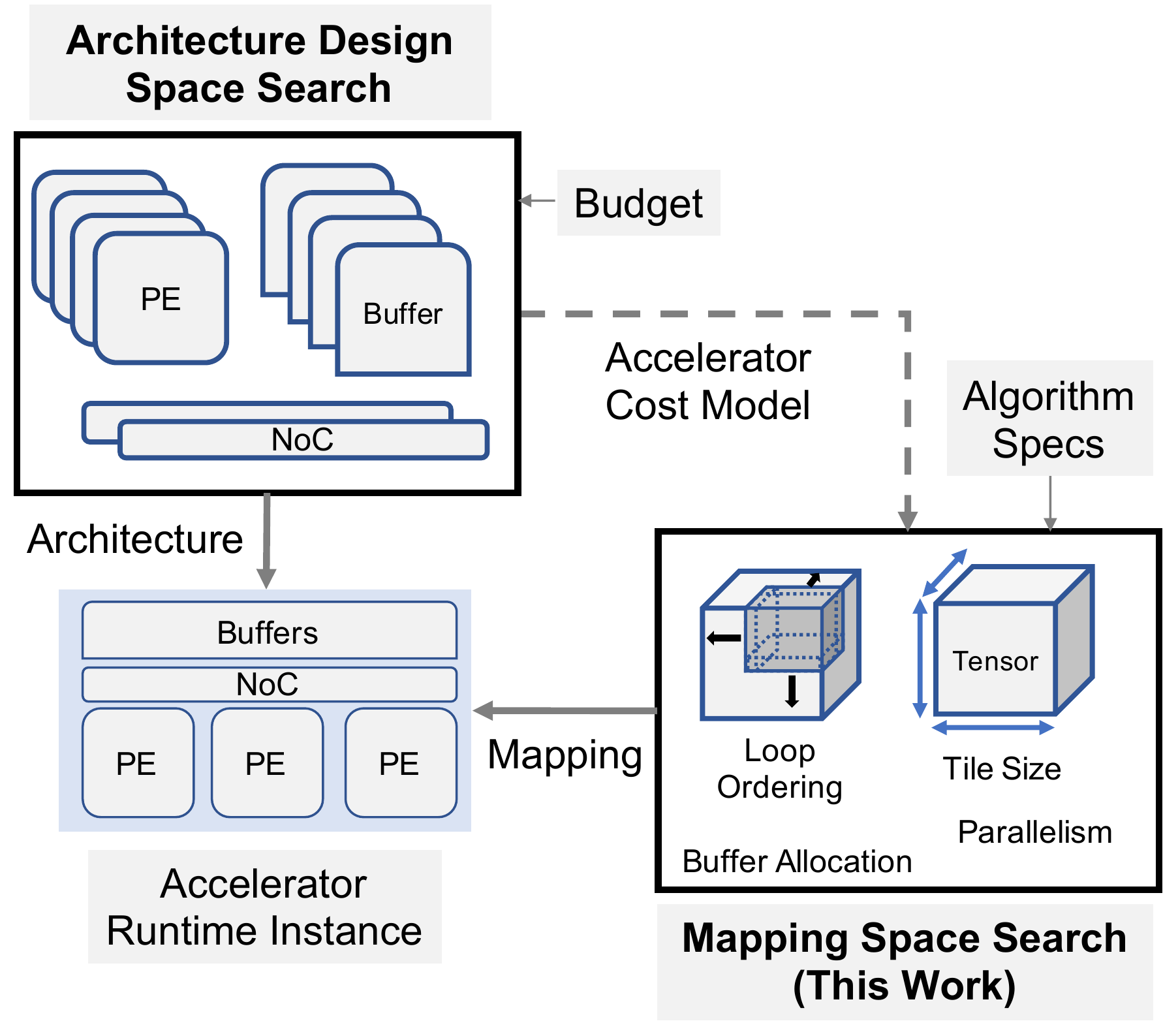}
  \caption{Architecture design and algorithm mapping in hardware accelerators.
  }
  \label{fig:design_process}
\end{centering}
\end{figure}

Although they are specialized, accelerators are often flexible~\cite{maeri, dna, morph, flexiflow, eyerissv2}, designed to support different parameterizations of a single algorithm to even a range of algorithms within or across domains.
This flexibility forces architects to decouple the act of designing the architecture from the act of mapping a specific \emph{problem}--a parameterized instance of an algorithm--onto the architecture.
This is shown in Figure~\ref{fig:design_process}.
First, pre-fabrication, architects 
choose architectural parameters to suit the budget and deployment requirements of expected target problems---a process referred to as \emph{Architecture Design Space Search}. 
This can include decisions such as the number of processing elements, on-chip buffer sizes, network-on-chip topology, bandwidth to off-chip memory, etc.
Second, post-fabrication and based on the designed-in flexibility of the hardware, 
architects or users map target algorithms to the hardware---referred to as \emph{Mapping Space Search}.
These decisions can include choosing how much of each buffer to allocate for each data structure,
mapping of computations to processing elements, etc., and are analogous to writing/compiling programs for general-purpose processors.

Mapping space search is an important problem, and currently faces severe scalability and performance challenges.
To start, prior work has shown that problem efficiency is very sensitive to the choice of mapping~\cite{eyeriss, scnn, morph, timeloop, maeri, flexiflow, dna}.
Further, the same studies illustrate how optimal mapping varies significantly depending on problem size and parameters (e.g., DNN model parameters), resource availability, performance and power requirements, etc.
This suggests that mapping space search will constitute an increasing recurring cost, as accelerators are re-targeted for new problems.

Making matters worse, simply gaining intuition for how to search through the map space, or how to pose the search to an automated tool, is an ad-hoc and expert-driven process.
Accelerators lack consistent hardware-software abstractions, such as instruction set architectures (ISAs) in the general-purpose computing world, and instead rely on bespoke configurable hardware components designed to provide higher degrees of control and efficiency.
Further, different accelerators tend to have different degrees of configurability in different hardware components, ranging from programmable networks on chip~\cite{maeri}, buffers~\cite{buffets,morph}, address generators~\cite{ganax}, etc.
While this may be suitable for experts with deep knowledge of both the architecture and algorithm, it clearly does not scale to non-experts programming new hardware with new algorithms.



Making matters even worse, while prior work has proposed tools and algorithms (i.e., \emph{Mappers}) for automatically searching the map space, all existing approaches have serious limitations due to search space complexity~\cite{morph,timeloop,tvm, release, flextensor}.
First, the search space is often high dimensional~(i.e., each degree of configurability induces a dimension), causing a combinatorial explosion of possible mappings and rendering exhaustive techniques ineffective~\cite{timeloop}.
Second, the search space is both non-convex (many local minima) and non-smooth (not differentiable), forcing prior work to rely on \emph{black-box optimization}~\cite{blackbox} approaches. 

To summarize, while configurable accelerators have demonstrated their potential, the lack of an efficient and high-quality Mapper hinders broader adoption.

\subsection{This Work}

This paper addresses the challenges above by proposing \emph{Mind Mappings}, a scalable and automated method to quickly and effectively perform mapping space search.

As mentioned previously, the key challenge hindering prior work is that mapping search space is non-smooth, forcing prior work to resort to black-box optimization techniques.
This is because the accelerator \emph{cost function}---which search algorithms use to evaluate the cost of a given candidate mapping---is non-smooth.
For example, the cost function might be an architectural simulator or the accelerator itself.

Mind Mappings addresses this challenge by constructing 
a \emph{differentiable} approximation of the cost function, called the \emph{surrogate}~\cite{surrogate_definition, surrogate_definition_0, surrogate_definition_1}.
Using the surrogate, Mind Mappings derives gradients for the cost function, with respect to candidate mappings, and uses those gradients to perform
a powerful first-order optimization technique, Gradient Descent~\cite{backprop, efficient_backprop}, to quickly find low-cost mappings.

The key insight here is that the differentiable surrogate of the actual non-differentiable cost function can provide us with approximate gradients, which are sufficient to guide the search along the direction of steepest descent, even in the absence of true gradients.
This insight simultaneously improves map space search quality and reduces map space search time, as gradients by definition point in the direction of the greatest reduction in cost.

Crucially, Mind Mappings formulates both predicting the cost of a mapping and finding the optimal mapping as \emph{learning} problems, thereby doing away with requiring expert knowledge in the target domain.
This paper makes the following contributions:
\begin{enumerate}
    \item  To the best of our knowledge, 
    our work is the first to enable 
    \textit{target domain-independent} mapping space search for programmable accelerators. 
    We require neither expert knowledge in the target application domain(s), nor any domain specific heuristics for handling programmable hardware attributes.
    \item To the best of our knowledge, 
    our work is the first to formulate mapping space search as a first-order optimization problem, enabling an efficient  gradient-based search that is able to quickly find high-quality mappings.
    \item 
    We extensively evaluate Mind Mappings across two target algorithms---CNNs and MTTKRP---comparing against multiple baseline search heuristics including simulated annealing (SA), genetic algorithms (GA), and reinforcement learning (RL).
    For CNNs and MTTKRP, the proposed search method finds mappings with an average $1.40\times$, $1.76\times$, and $1.29\times$ (when run for a fixed number of steps) and $3.16\times$, $4.19\times$, and $2.90\times$ (when run for  a fixed amount of time) better energy-delay product over SA, GA, and RL, respectively. 
    \item To facilitate further adoption, we provide a reference implementation of the Mind Mappings framework here: \url{https://github.com/kartik-hegde/mindmappings}.
\end{enumerate}

%% file: ASPLOS CR/mapspace.tex
\section{Background}
\label{sec:mapspace}



In this section, we formally define the algorithm-accelerator mapping space search problem and elaborate with an example.

\subsection{Algorithm-Accelerator Mapping Space}
\label{subsec:mapspace_search}


In this paper, we assume that a hardware accelerator $a$ and a target problem $p$ is given, where a \emph{problem} is a parameterized instance of an algorithm.
For example, if matrix multiplication is the target algorithm, an example target problem is a matrix multiplication between two matrices of fixed shape (dimensions), 
irrespective of the contents of the matrices.
We begin by defining a mapping $m$ and the mapping space $M$.
\begin{definition}
\label{def:mapping}
\emph{Mapping.} A mapping $m \in P_0\times\dots\times P_{D-1}$ is a $D$-tuple, where $D$ is the number of programmable attributes of the given accelerator $a$.
Each element in the mapping vector, $m_d$ for $d\in[0, D)$ belongs to the domain of the $d$-th programmable attribute, $P_d$.
\end{definition}

\begin{definition}
\label{def:mapspace}
\emph{Mapping Space.} Given an \emph{accelerator} $a$ and a target \emph{problem} $p$,
we define a mapping space as
$$M_{a,p} = \{m \in P_0\times\dots\times P_{D-1}\;|\;a(m,i)==p(i)\;\forall i\in I_p\}$$ 
where $I_p$ denotes the possible inputs~(e.g., all possible matrices with a specific shape) to a problem $p$, $a(m,i)$ 
denotes the accelerator's output given mapping $m$ and input $i$, and $p(i)$ denotes the golden reference output given input $i$ for problem $p$.
\end{definition}

In other words, $M_{a,p}$ is the set of mappings that result in functional correctness for the given problem $p$ on the accelerator $a$.
We call such mappings \emph{valid} for $a$ and $p$ and write $M_{a,p}$ as $M$ for short when the context is clear.
Intuitively, the different $m\in M$ can be thought of as different ``programs'' representing problem $p$ on accelerator $a$.
An example of a mapping space is given in Section~\ref{subsec:mapspace_example}.

With this in mind, the size of the mapping space varies based on the programmability of the underlying accelerator and the problem $p$.
On one extreme, the size of $M$ (denoted $|M|$) can be 1 for a fixed-function ASIC designed to execute one fixed problem.
In general, $|M|=O(\prod_{d\in[0,D)} |P_d|)$, where the number of attributes $D$ and the size of each attribute space $|P_d|$ is large.
The Big-Oh captures how some mappings in the Cartesian product of assignments to programmable attributes may be invalid.

\subsection{Mapping Space Search}
\label{subsec:map_space}

\emph{Mapping Space Search}
is the combinatorial search problem to find the mapping $m_{opt} \in M$ that minimizes the \emph{cost} $f$, where the cost function $f$ is the optimization objective set by the designer~(discussed further in Section~\ref{subsec:cost}).
That is,
\begin{equation}
m_{opt} = \argmin_{m\in M_{a, p}} f(a,m)
\label{eq:schedule}
\end{equation}

where the user specifies the problem $p$ and architecture $a$. 
We denote $f(a,m)$ 
as $f(m)$ for short. 
In theory, mapping space search can be performed at compile time or at run time. 
In practice, programmable accelerators today either perform the search completely offline, e.g., when compiling a new problem to the architecture~\cite{morph,maestro,scalesim}, or partly offline/partly online~\cite{nvdla}.

In this paper, we assume $f$ is a function of $a$ and $p$---not the problem input $i$.
This holds for several important workloads, e.g., kernels in dense tensor algebra such as dense matrix multiplication and deep neural network training/inference~\cite{timeloop}.
However, it does not hold when input-dependent optimizations, e.g., data sparsity, influence cost~\cite{extensor}.
We consider efficient mapping space search for input-dependent mappings to be important future work.

\subsection{Cost Function}
\label{subsec:cost}
The cost function $f$ (Equation~\ref{eq:schedule}) estimates the cost of running the given mapping $m$ on the given hardware accelerator $a$.
This serves as the objective for optimization in the mapping space search.
It is up to the designer to formulate the cost function based on the design criteria.
For example, the cost function can be formulated as a weighted sum or product of several factors, or as a prioritized order of measurable metrics~\cite{timeloop}.
For example, $f(a,m) = \sum_{k=0}^{K-1} w_k f_k(a, m)$
where $K$ is the set of all the factors considered by the designer and $w_k$ is the importance assigned to the $k$-th factor.
Factors can include various measures such as power, performance, or meta-statistics such as the number of buffer accesses, etc., and designers may choose to assign appropriate weights for each of the costs based on the requirements/factors.
For example, if $f_k$ represents the number of DRAM accesses, $w_k$ might represent the energy per DRAM access.
Importantly, the function computing each factor $f_k$ need not be smooth, differentiable, etc.

\section{Example Mapping Space: 1D-Conv}
\label{subsec:mapspace_example}

\newcommand{\inp}{\mathsf{I}}
\newcommand{\out}{\mathsf{O}}
\newcommand{\filt}{\mathsf{F}}

We now describe the mapping space for a hardware accelerator designed to perform 1D-Convolution (1D-Conv) with energy-delay product as the cost function.
This represents a simplified version of the accelerator and algorithm (Convolutional Neural Nets/CNNs) that we evaluate in Section~\ref{sec:eval}, and we will refer to the example in Section~\ref{sec:method} to explain ideas.

Algorithmically, for filter $\filt$, input $\inp$, and output $\out$, 1D-Conv is given as
\begin{align}
&\out[x] = 
    \sum_{r=0}^{R-1} 
        \inp[x+r] * \filt[r] \label{eqn:1dconv} \\
    \nonumber &0 \le x < W-R+1
\end{align}
for input width $W$ and filter size $R$.
Using the terminology in Section~\ref{subsec:mapspace_search}, the 1D-Conv algorithm forms a family of problems, where each problem $p$ corresponds to a specific setting of $W$ and $R$.

1D-Conv in Equation~\ref{eqn:1dconv} can be represented as a loop nest:
\begin{lstlisting}[language=C,caption={\small Untiled 1D-Convolution.},captionpos=b,label={code:untiled}]
for(x=0; x<W-R+1; x++) {
	for(r=0; r<R; r++) {
		O[x] += I[x+r] * F[r]; } }
\end{lstlisting}
We represent the ordering of the loops in the above loop nest as $W\rightarrow R$, meaning iteration over $W$ and $R$ is the outer and inner loop, respectively.
Note that due to the commutativity of addition~(ignoring floating point errors), we can freely interchange the loops, i.e., as $R\rightarrow W$.

We can also add additional levels to the loop to model tiling or blocking.
For example, if $\filt$ cannot fit in a buffer, we can tile it as shown here:
\begin{lstlisting}[language=C,caption={\small Tiled 1D-Convolution.},captionpos=b,label={code:tiled}]
for(rc=0; rc<Rc; rc++) { // Rc=ceil(R/Rt);
  for(x=0; x<W-R+1; x++) {
    for(rt=0; rt<Rt; rt++) {
      roff = rc*Rt + rt;
      O[x] += I[x+roff] * F[roff]; } } }
\end{lstlisting}
That is, we complete the 1D-Conv for an $R_t$ chunk of $\filt$ at a time, adding an outer loop over the number of tiles $R_c$.
This loop nest is therefore written as $R_c\rightarrow W\rightarrow R_t$. 
Beyond tiling, we can describe parallelism in the computation by conceptually pre-pending \texttt{parallel} before a given loop(s), indicating that iterations of that loop can run in parallel.

\begin{figure}[t]
\begin{centering}
  \includegraphics[width=0.70\columnwidth]{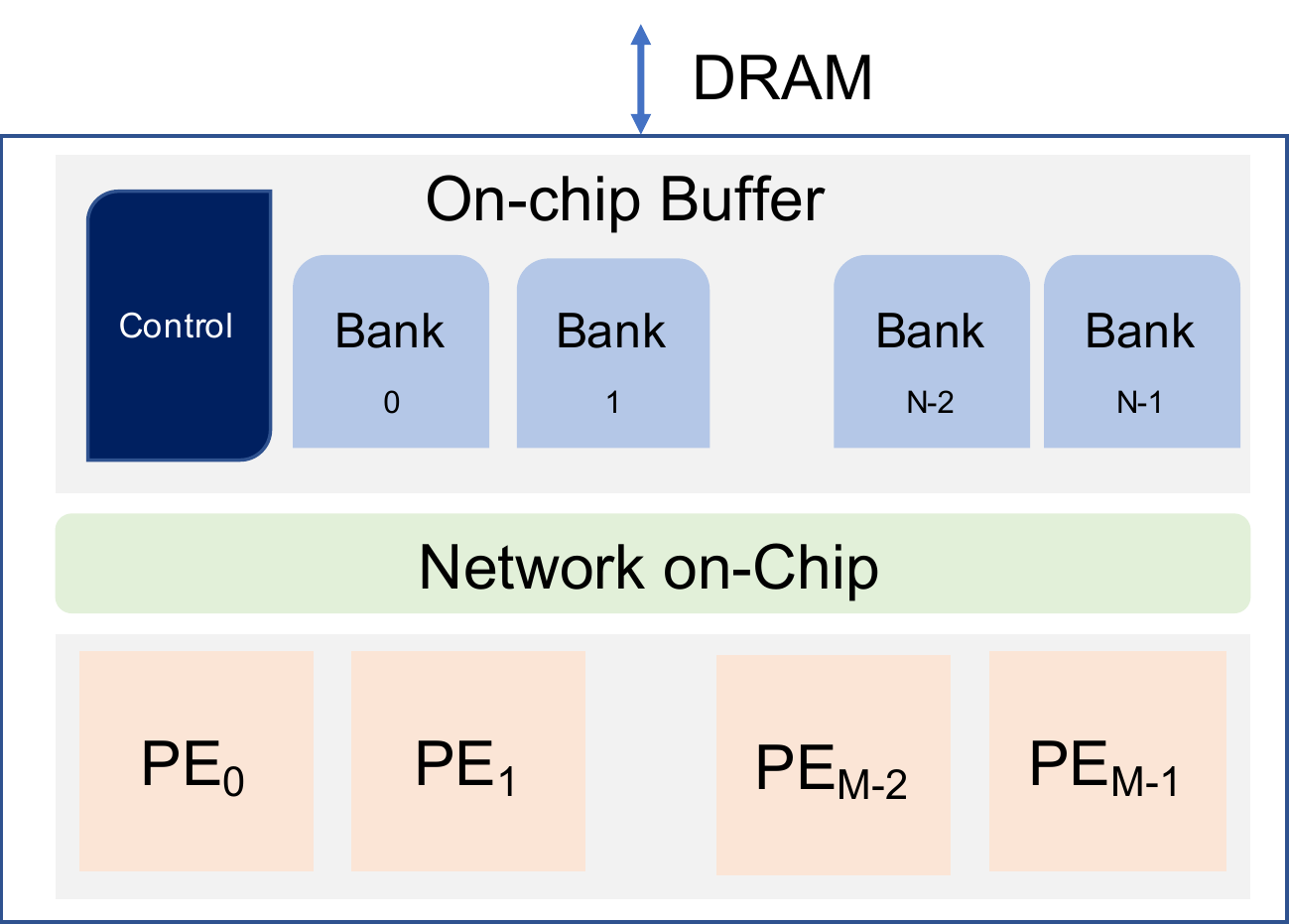}
  \caption{A basic hardware accelerator with $M$ processing elements, on-chip buffer with $N$ banks, and a NoC.}
  \label{fig:accelerator}
\end{centering}
\end{figure}

An example target accelerator for the 1D-Conv algorithm is depicted in Figure~\ref{fig:accelerator}.
At a high-level, it has $M$ processing elements, an on-chip buffer with $N$ banks allocatable to different tensors at bank granularity, a flexible NoC that can be optimized for different communication patterns such as broadcast, unicast, etc.

Let us assume that the accelerator's programmable attributes are:
\begin{enumerate}
    \item $P_0 = \mathbb{R}^3$: a 3-tuple indicating the percentage of banks allocated to each of $\inp$, $\out$, and $\filt$.
    \item $P_1 = \mathbb{Z}^{3}+$: a 3-tuple representing the tile shape of $\inp$, $\out$, and $\filt$ to be fetched from DRAM.\footnote{Note that in 1D-Conv, tiles are 1-dimensional.  Hence, shape is representable as a scalar.}  The $+$ is to facilitate multiple levels of tiling, if applicable.
    \item $P_2 = \{W\rightarrow R, R\rightarrow W\}$: loop order for the untiled 1D-Conv algorithm represented in Code~\ref{code:untiled}. 
    To support more loops levels due to tiling (as in Code~\ref{code:tiled}), we add additional tiled loop orders to $P_2$ (e.g., $R_c\rightarrow W\rightarrow R_t$).
    
    \item $P_3 = \mathbb{Z}+$: the loop bound for each loop, e.g., $R_c, W+R-1, R_t$ in Code~\ref{code:tiled}.
    Note, we write this attribute explicitly for clarity.  In this example, loop bound can be inferred from the current assignment to $P_1$ and $P_2$. 
    \item $P_4 =  \{\mathrm{unicast},\;\mathrm{multicast},\mathrm{broadcast}\}^3$: NoC communicating patterns for each of the 3 tensors.
    \item$P_5 = \mathbb{Z}+$: Amount of parallelism per PE for each loop level.
\end{enumerate}

\textbf{Mapping Space.} 
Given the above, one can construct a mapping space $M_{a,p}$ for the accelerator and specific 1D-Conv problem (i.e., the specific setting of $W$ and $R$).
The accelerator has $D=6$ programmable attributes and the mapping space size is bounded by the size of the Cartesian product of assignments to each programmable attribute.
As discussed in Section~\ref{subsec:mapspace_search}, the mapping space size will be smaller than this upper bound, as some complete assignments to programmable attributes are invalid, e.g., $R_t < R$ must hold since $R_t$ is a tile of $R$. 

\textbf{Cost Function.} In the above example, the cost of a mapping $f(m)$ is defined as $\mathrm{energy}*\mathrm{delay}$.
This can be obtained by simulators or analytical models that represent the actual hardware or using the actual hardware itself.
For example, Timeloop~\cite{timeloop} (which we use in our evaluation) is a tool that can calculate mapping cost for algorithms representable as affine loop nests (e.g., convolutions).

\subsection{Challenges in Mapping Space Search} 
\label{subsec:challenges}

\begin{figure}[h]
\begin{centering}
  \includegraphics[width=0.95\columnwidth]{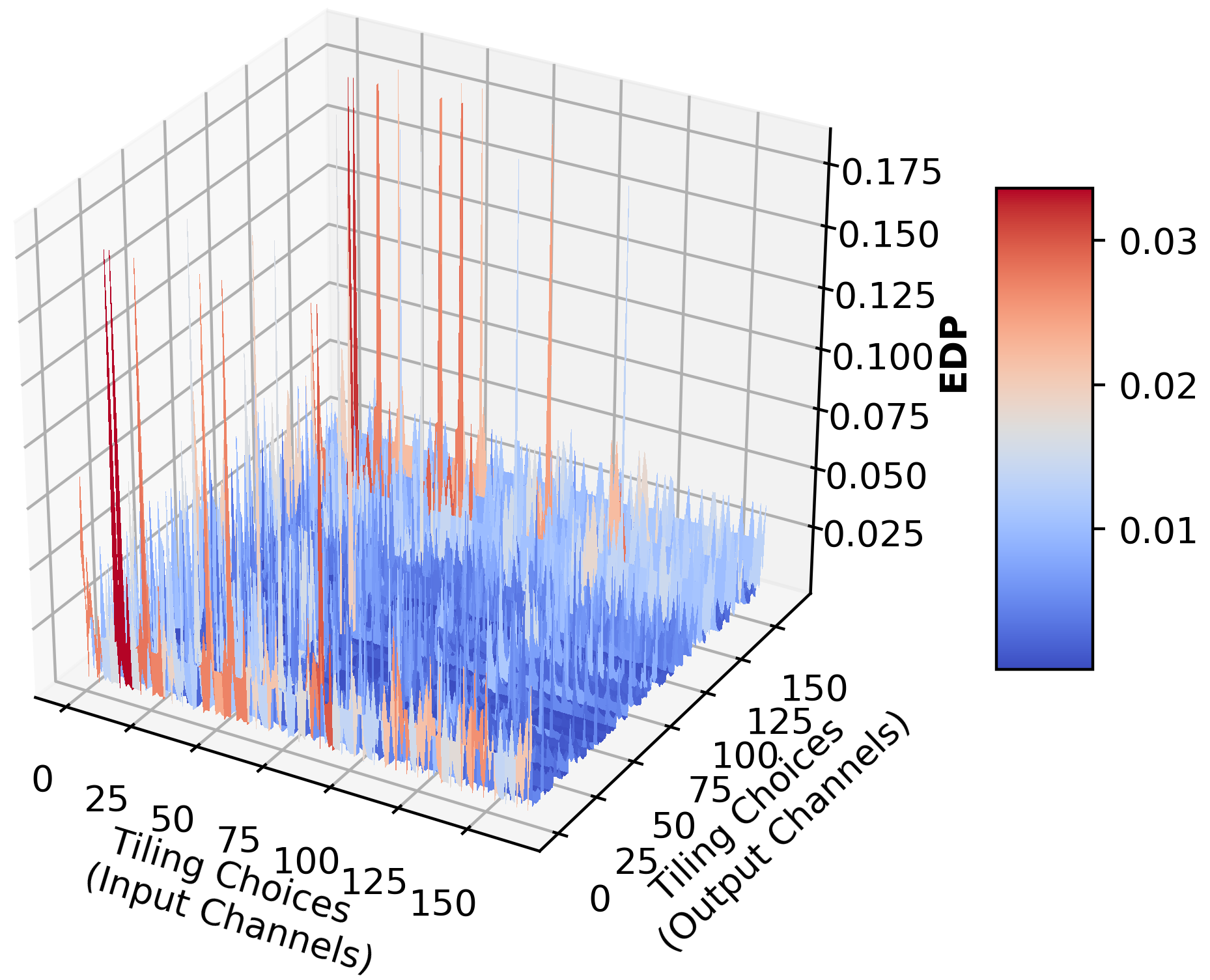}
  \caption{Cost surface plot for the accelerator we evaluate in Section~\ref{sec:eval} for CNNs. Darker red indicates higher EDP, and darker blue indicates lower EDP. Mind Mappings approximates this non-smooth surface with a differentiable surrogate to enable gradient-based optimization.
  }
  \label{fig:error_surface}
\end{centering}
\end{figure}

A major challenge in mapping space search is the nature of cost function $f$, which is non-convex and non-smooth.
For example, consider the 1D-Conv example from the previous sub-section.
The cost of a mapping $f(m)$ is influenced by the various programmable attributes $P_d$ in subtle ways.
For example, consider $P_0$, the attribute that represents buffer allocation for each operand/result tensor $\inp$, $\out$, and $\filt$.
If the $\filt$ tensor is 1~KB in size and allocation falls short of 1~KB, the mapping cost will see a significant energy bump as valid mappings will be forced to tile $\filt$, requiring some operand(s) to be re-fetched multiple times to fully compute $\out$.
In other words, seemingly minor changes to mapping $m$ can result in non-smooth changes to the overall cost $f(m)$.

To illustrate this, Figure~\ref{fig:error_surface} plots the cost surface
for the programmable accelerator running Convolutional Neural Network (CNN) layers that we evaluate in Section~\ref{sec:eval}.
In the figure, the $x$- and $y$-axis represent different choices of tile sizes for two different input tensors, 
while the $z$-axis represents the cost $f$ in terms of the energy-delay product (EDP).
Evident from the plot, the search space is spiky and non-smooth in nature.
Due to this, obtaining useful statistics such as the gradients~(first-order), Hessians~(second-order) of the search space is not possible, requiring the search for optimal mapping~(Equation~\ref{eq:schedule}) to use black-box optimization approaches such as Simulated Annealing~\cite{simanneal}, Genetic Algorithms~\cite{genetic}, etc.
Making matters worse, the search space is clearly non-convex, i.e., many local minima, making the search even harder.
Given the humongous search space size~($\approx10^{25}$ in this example), black-box optimization approaches struggle to find high quality mappings in few iterations.

%% file: ASPLOS CR/method.tex
\section{Method}
\label{sec:method}







\begin{figure*}[t]
\begin{centering}
  \includegraphics[width=\textwidth]{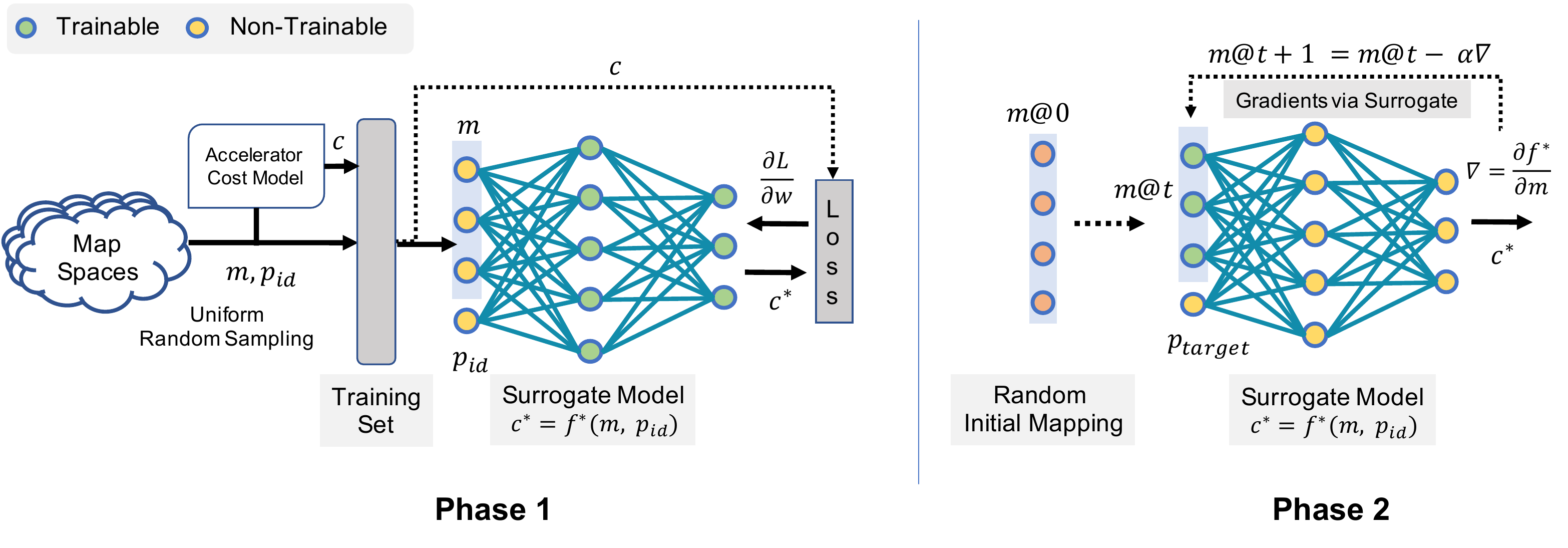}
  \caption{Mind Mappings search procedure. \textbf{Phase 1:} Training the surrogate model $c^*=f^*(m, p_{id})$ based on (mapping, problem id, cost) tuples ($m$, $p_{id}$, $c$).  DNN (surrogate) weights $w$ are trained with back-propagation. \textbf{Phase 2:} Given a target problem $p_{target}$, use the trained surrogate model to iteratively guide a random initial mapping $m @ 0$~(``mapping at search iteration 0'') towards an optimal mapping $m_{opt}$. In each iteration, $m @ t$ is updated using back-propagation with a gradient $\nabla$ of $f^*$ based on $m @ t$ with a learning rate $\alpha$. The trained model weights $w$ and the target problem $p_{target}$ are held constant in this phase.
}
  \label{fig:training}
\end{centering}
\end{figure*}




Due to the non-smooth nature of the cost function described in the previous sections, designers are forced to use black-box optimization approaches to find optimal mappings.
By definition of being black box, they cannot take advantage of structure within the accelerator cost function, which puts them at a disadvantage in effective mapping space search.
Mind Mappings circumvents this by \emph{approximating} the search space as a smooth, differentiable function.
This turns mapping space search into a white-box optimization problem, and enables gradient generation to improve the search.

Mind Mappings is a two-phase procedure, as shown in Figure~\ref{fig:training}.
We start with a target algorithm.
The goal is to find low-cost mappings for a potentially unbounded number of target problems given that algorithm.
To accomplish this:
First (Phase 1, Section~\ref{subsec:func_approx}), we train a \emph{differentiable} surrogate model to approximate the accelerator's cost function for all problems making up the target algorithm.
Second (Phase 2, Section~\ref{sec:design:mapping_search}), we perform Gradient Descent on the surrogate model to generate low-cost mappings for the target problems. 
Phase 1 and 2 are performed offline and online, respectively.


To amortize surrogate training cost (Phase 1), 
we train the surrogate to generalize to unseen problems and 
reuse the surrogate across the potentially many target problems in Phase 2.
For example, given 1D-Conv from Section~\ref{subsec:mapspace_example}, Phase 1 trains the surrogate on mappings corresponding to representative $W$ and $R$ values, so that the surrogate will later be able to interpolate and return accurate costs for mappings belonging to \emph{unseen} $W$ and $R$ values.
Then, Phase 2 uses the surrogate to search for low-cost mappings for those unseen $W$ and $R$ values.
That is, the surrogate is trained once, offline per target algorithm.





We now discuss Phase 1 and 2 in more detail.
We rely on the 1D-Conv example from Section~\ref{subsec:mapspace_example} to explain ideas.


\subsection{Phase 1: Approximating the Map Search Space}
\label{subsec:func_approx}


As discussed in Section~\ref{subsec:map_space}, the cost function $f$ that maps input mappings to a designer-defined cost is non-smooth and non-differentiable.
To make $f$ differentiable, which will allow us to generate gradients, we use \emph{function approximation}~(FA) with differentiable surrogates.
FAs are often used to reduce a function's dimensionality, which has been shown to simplify optimization problems in different fields such as reinforcement learning~\cite{func_approx_0,func_approx_1}.
It has also been used in prior works~\cite{surrogate_0,surrogate_1, ithemal, tvm} to improve the speed of evaluation of $f$ to enable rapid searches.

With FA, we generate a smooth, differentiable approximation of the cost function, denoted \fapprox, called the \emph{surrogate}.
Given a mapping $m$ and problem $p$, the surrogate predicts the cost $c^* =$ \fapprox$(m,p)$, where $c^*$ approximates the actual cost, $c=f(m)$. Therefore, each surrogate is specialized for a given accelerator and the target algorithm, but should be able to return accurate costs for the different problems making up the algorithm.
Notice, $p$ is specified as an input to \fapprox.
This is a notational convenience that will be important in Phase 2.

While there are multiple choices of differentiable surrogate functions, we use Multi-layer Perceptron~(MLP)-based Deep Neural Networks~(DNNs) in this paper as 
they are known to be capable of modeling high-dimensional functions 
and feature a mature training infrastructure due to the recent advances in Deep Learning.
We leave the question of whether simpler, differentiable models are sufficient as future work.

Figure~\ref{fig:training}, Phase 1, gives an overview of how to construct/train the surrogate. 
At a high level:
We construct a training dataset of input mappings $m$ and their associated costs $c=f(m)$, where
$f$ is the accelerator's reference cost model.
Each element in the training set is stored as a 3-tuple: mapping $m$, problem identifier $p_{id}$ (from which $m$ was sampled) and reference model cost $c$.

The distance between the predicted cost, \capprox, and the actual cost $c$ is used to generate a loss, which is used to train the surrogate using back-propagation~\cite{backprop}.
The training procedure can be carried out until satisfactory loss is reached, and well-studied techniques in deep learning can be used to improve the speed of convergence~\cite{dropout, Imagenet}.

Superficially, the above process is a ``standard'' supervised training with Stochastic Gradient Descent~(SGD), widely used in modern deep learning approaches.
Getting it to work properly for mapping space search, however, entails addressing multiple issues, as described below.

\subsubsection{Generating the Surrogate Model Training Set}
\label{sec:method:gen_train}

The first step is to build a training set to train the surrogate model.
We face the following questions:


\textbf{(1) Which map spaces should be used to populate the training set?}
The naive approach is to populate the training set with mappings associated with a single problem for the target algorithm.
This approach fails because the resulting surrogate will not generalize to unseen problems, requiring us to re-train the surrogate for each new problem we encounter.
For example, a new surrogate will be needed every time the $W,R$ settings change for 1D-Conv---clearly undesirable.
Instead, we generate training points by uniformly sampling from multiple map spaces, thereby generalizing the surrogate across the family of problems associated with the target algorithm.
We empirically observe that the model is able to interpolate and predict correctly for problem instances it hasn't seen before.


\textbf{(2) Based on the choice of map spaces, which mappings should we sample to populate the training set?}
There are three issues here. First, we must be able to check if a mapping is valid, i.e., belongs to a specific map space.
For this, we assume there exists a membership testing function $\mathsf{isMember}(m,p)$ for every accelerator $a$ which returns true if $m\in M_{a,p}$ and false otherwise.

Second, we must decide whether to populate the training set with only valid members of each map space---i.e., mappings that are functionally correct for their given problem---or also consider invalid mappings.
In this work, we only populate the training set with valid mappings.
Considering invalid mappings may enable the surrogate 
to better avoid functionally incorrect mappings in Phase 2, e.g., by assigning them infinite cost.
However, this may face implementation challenges such as exploding gradients and slow convergence~\cite{exploding_grad}.

Third, we need to decide on a sampling strategy to sample from each map space.
One option, which we use in this work, is to sample uniformly at random. 
Specifically, given a problem $p$, we sample from the associated map space $M_{a,p}$ uniformly to generate a mapping and re-sample if the mapping is not valid (see above).
This ensures a reasonable representation of the map space, but might suffer from under-sampling from regions that are more important from a training perspective.
Other advanced sampling methods can be used, such as sampling from a probability distribution trained to maximize the amount of learning per sample~\cite{gumbel}.
As seen from our evaluation~(Section \ref{sec:eval}), uniform random sampling facilitates training well and therefore we leave improved sampling methods to future work.

\textbf{(3) How to uniquely associate each mapping $m$ with its map space $M_{a,p}?$}
It is possible to have the same mapping $m$ present in multiple map spaces, where each mapping instance has a different cost $c$.
Therefore, to ensure correct generalization of the surrogate across different problems for the given algorithm,
we must uniquely identify each mapping in the training set with its map space.
For this, we need to tag each mapping $m$, added to the training set, with a problem identifier $p_{id}$, unique to the map space associated with its problem $p$.
In this paper, we encode each $p_{id}$ as the specific parameterization of the problem, e.g., a tuple indicating the $W,R$ values associated with the problem for the 1D-Conv algorithm~(Section~\ref{subsec:mapspace_example}).

\textbf{(4) How to calculate cost per mapping?}
To train the surrogate, we require a reference cost $c=f(m)$ that can be used to calculate the loss w.r.t. surrogate's predicted cost \capprox. 
This can be estimated via running the problem with mapping $m$ on the target hardware or using a cost function estimator such as those described in Section~\ref{subsec:cost}.
As this is a one-time, offline procedure that generalizes over different problems for the target algorithm, its cost is amortized over multiple mapping space searches performed using this surrogate.

We now describe input mapping vector and output cost representation.
We describe ideas and challenges here.
The concrete representations we use for our evaluation are detailed in Section~\ref{subsec:eval:sensitivity}.

\subsubsection{Input Mapping Representation}
\label{sec:method:input_rep}

As described in Section~\ref{subsec:mapspace_search}, the mapping vector $m$ is a $D$-tuple consisting of the accelerator's programmable attributes, which needs to be converted to a representation that can be used to train the surrogate.
There are two issues here.
First, while each programmable attribute $P_d$ can have different representations, e.g., vector, integer, float, boolean, etc., the input to the surrogate needs to be a single vector of floats.
We resolve this by converting each attribute to a scalar or a vector of floats, flattening multiple vectors into a final mapping vector as needed.
For example, $P_4$ in 1D-Conv (Section~\ref{subsec:mapspace_example}) has 3 discrete choices, which can be converted to a vector of 3 floats which are one-hot encoded.
The choice of float for the mapping vector datatype isn't fundamental; in what follows, we refer to each float/element in the mapping vector as a value.


Second, in some cases, it may be desirable to have variable-length mapping vectors for different problems~(e.g., to encode variable levels of tiling), but the input vector to the surrogate is often fixed~(e.g., as in a Multi-layer Perceptron).
While we deal with fixed-dimensionality mappings in this work, the above can be easily handled via embeddings~\cite{embedding}, a widely used method to deal with different input lengths in areas such as Natural Language Processing and Recommendation Systems.

Finally, we normalize each value in each mapping to have mean 0, standard deviation 1---in a process akin to input whitening~\cite{Alexnet, Imagenet}. 
That is, let $m'^{i}_{d}$ be the $d$-th value in the $i$-th mapping in the training set, where $m'$ means the mapping $m$ has been flattened into a vector of $D'$ values as described above.
Then, we normalize each $m'^{i}_{d}$ with respect to other $m'^{j}_{d}$ in the dataset.

\subsubsection{Output Cost Representation}
\label{sec:method:output_rep}


A crucial decision to make is to choose a representation for the cost vectors $c$ and $c*$ that facilitates a high quality surrogate.
A straightforward approach, often used by prior works, is to represent costs as a combined statistic or the final target, such as EDP, performance/watt, etc.

Instead, we encode costs as a vector of \emph{meta-statistics} as we observed that this results in higher quality surrogates.
For example, for the surrogate we evaluate in Section~\ref{sec:eval}, we represent cost as a vector containing the energy spent accessing each level of the memory hierarchy by each data type (e.g., input/output tensor in 1D-Conv), compute utilization, total cycles, and total energy, although the final metric of interest is EDP.
We empirically found that this rich output representation enabled the surrogate model to achieve a $32.8\times$ lower mean-square error to the ground truth EDP, relative to surrogates that output EDP directly.

Additionally, we normalize the output cost vector with respect to a theoretical lower bound cost for the target problem, to reduce the variance in output values.
We use a conservative lower bound that assumes perfect data reuse and perfect compute utilization.
For example, for 1D-Conv from Section~\ref{subsec:mapspace_example}, the lower bound for cycles is given by $((W-R+1)*R)/max\_flops$ (assuming 1 cycle/FLOP), whereas the lower bound for energy is given by 
$(W + W-R+1 + R)$ times the energy needed to access each word of data once.
We evaluate an architecture with an inclusive buffer hierarchy, meaning that the energy needed to access each word once the sum of the energies per access for each level of the memory hierarchy.
We note that while this lower bound is possibly not achievable, it is only meant to give us a good normalization metric.

Finally, similar to inputs~(Section~\ref{sec:method:input_rep}), each value in the output vector is normalized to have mean 0 and standard deviation of 1 with respect to the corresponding values in other cost vectors in the training set.

\subsection{Phase 2: Gradient Search to Find High-Quality Mappings}
\label{sec:design:mapping_search}

Phase 2, the online part of the search procedure, finds a low-cost mapping $m_{opt}$ for the target problem $p_{target}$, as depicted in Figure~\ref{fig:training}.
We achieve this by leveraging the differentiability of the surrogate model \fapprox~(Phase 1) 
to obtain gradients, where gradients represent the change in $m$ that maximally reduces the cost $f(m)$.
With access to gradients, unlike black-box optimization approaches, we can perform powerful first-order optimization methods which, in effect, incrementally guide any random valid mapping $m$ towards $m_{opt}$ using Gradient Descent.
The key insight here is that, while the cost function representing the accelerator itself is non-differentiable, its differentiable approximation \fapprox \ should give us access to approximate gradients that can guide the search along the direction of steepest descent.




\textbf{Gradient Descent with MLP-based Differentiable Surrogates.}
Section~\ref{subsec:func_approx} described the procedure to obtain a differentiable surrogate \fapprox \ by training an MLP to approximate the cost function.
We now use the surrogate to generate gradients that indicate the direction of steepest descent with respect to a \emph{candidate mapping} $m$, i.e.,  $\nabla $\fapprox$_{p_{id}}(m) = [\partial $\fapprox$/\partial m_0, .. , \partial $\fapprox$/\partial m_{D-1}]$. 
$\nabla $\fapprox$_{p_{id}}$ is computed using the chain rule across layers of the MLP (surrogate), assuming the MLP parameters and $p_{id}$ are fixed.

We use the generated gradients to perform a standard Gradient Descent starting from a randomly-chosen initial mapping $m@0$ (``$m$ at iteration 0''), setting $p_{id}=p_{target}$.
That is, we iteratively compute $\nabla $\fapprox$_{p_{target}}(m@t)$ and combine it with $m@t$ to form the next candidate mapping $m@{t+1}$.
This process is shown in more detail in Figure~\ref{fig:training}.


\textbf{Projected Gradient Descent.} 
Applying Gradient Descent 
to mapping space search presents two issues.
First, after applying gradients, each value $m_d$ in the mapping 
vector falls potentially outside the domain of programmable attribute $P_d$.
To address this, we round each $m_d$ to the nearest value in $P_d$.
Second, after rounding each $m_d$, the overall $m$ may be invalid with respect to $p_{target}$, i.e., $m \centernot\in M_{a,p_{target}}$.
To ensure the final mapping is valid, we check mapping validity of $m@t$ at each step $t$.
If validity fails, we calculate nearest neighbor valid mappings based on euclidean distance to $m@t$ and switch $m@t$ to the nearest valid mapping before continuing the search.
This is a standard approach, often referred to as \emph{Projected Gradient Descent}~\cite{projected_gradient_descent}, used in applications where gradient may steer the parameters out of the valid region~\cite{projected_0, projected_1}.

\textbf{Avoiding Local Minimas.}
\label{subsec:local_minima}
Gradient Descent is infamous for getting stuck in local minima for non-convex optimization problems~\cite{local_minima} and our scheme runs the same risk.
To handle non-convexity, we introduce randomness at regular intervals throughout the search.
Specifically, we add an outer loop to the Gradient Descent algorithm described above, where after every $N$ iterations, a new random mapping vector is introduced.
The decision to replace the current mapping $m@t$ with the newly sampled valid mapping is based on a probability function $\mathsf{accept}$
that helps us balance the trade-off between exploration and exploitation.
The choice of $N$ and the probability function is up to the implementation.
For example, the designer may choose to gradually
decay the probability of accepting a mapping with higher cost than the already-seen mappings over time, similar to Simulated Annealing.
We follow this strategy in Section~\ref{sec:eval}.

Overall, Phase 2 performs the following steps until termination, given a target problem $p_{target}$:
\begin{enumerate}
\item Choose a random valid mapping vector $m@t$ where $t=0$.
    \label{phase2:step1}
\item Compute $c^*=$\fapprox$(m@t, p_{target})$ by forward propagation through the MLP.
\label{phase2:step2}
\item Derive gradients using back-propagation via the surrogate with respect to $m@t$, $\nabla = \partial $\fapprox$/\partial m@t$.
\item Update the mapping vector as $m@{t+1} = m@t - \alpha \nabla$, where $\alpha$ is a user-specified learning
rate.
\item Project $m@{t+1}$ to the valid target map space.
\item If $t\%N==0$, sample a random valid mapping $m_{rand}$. If $\mathsf{accept}(m_{rand}, m@t, T)$ returns true, update $m@{t+1}=m_{rand}$, where $T$ is a temperature term that is decayed over time.
\item Increment $t$ and go to Step~\ref{phase2:step2} until completion.
\end{enumerate}

%% file: ASPLOS CR/evaluation.tex
\section{Evaluation}
\label{sec:eval}

We now evaluate Mind Mappings.
We design representative flexible hardware accelerators using Timeloop~\cite{timeloop} and search through their map spaces in the context of two target algorithms, while comparing against several other search methods.


\subsection{Experimental Setup}

\subsubsection{Algorithms}

We evaluate Mind Mappings by evaluating 
two target algorithms, Convolutional Neural Networks~(CNN) and Matricized tensor times Khatri-Rao product~(MTTKRP).
We evaluate two target algorithms to demonstrate generality, and selected these two in particular given the ongoing effort in the architecture community to build efficient hardware accelerators for CNNs~\cite{eyeriss, DaDianNao, shidiannao, scalpel, scnn, cnvlutin, ucnn} and MTTKRP~\cite{extensor,tensaurus}.
CNN layers feature similar, but higher-dimensional, computations as our 1D-Conv example from Section~\ref{subsec:mapspace_example}.

\textbf{CNN-Layer.} 
CNNs have seen widespread success in modern Deep Learning applications such as image recognition, video analytics etc. 
A CNN layer takes $N$ 2D images of resolution $W \times H$ with $C$ channels and $K$ filters of resolution $R \times S$ and produces an output of size $X \times Y$ with $K$ channels.
Value of $X$ and $Y$ can be calculated 
from $W$ and $H$ respectively as $(W-R+1)/stride$ and $(H-S+1)/stride$, respectively.
Mathematically, a CNN Layer 
is given by Equation~\ref{eq:conv2d}.

\begin{align}
&\out[(k, x, y)] = 
    \sum_{c=0}^{C-1} \sum_{r=0}^{R-1} \sum_{s=0}^{S-1} 
        \filt[(k, c, r, s)] * \inp[(c, x+r, y+s)] \label{eq:conv2d} \\
    \nonumber &0 \le k < K, 0 \le x < W-R+1, 0 \le y < H-S+1
\end{align}

\textbf{MTTKRP.} MTTKRP~\cite{mttkrp} is a key kernel in Tensor Algebra that is a bottleneck in applications such as tensor decompositions~\cite{kolda_survey}, alternating least squares~\cite{als}, Jacobian estimation~\cite{mttkrp_use}, etc.
MTTKRP takes a 3D tensor $A$, and matrices $B$ \& $C$ to produce and output matrix by contracting across two dimensions, as described in Equation~\ref{eq:mttkrp}.

\vspace{-3mm}
\begin{align}
&\out[(i,j)] = 
    \sum_{k=0}^{K} \sum_{l=0}^{L}
        A[(i,k,l)] * B[(k,j)] * C[(l,j)] \label{eq:mttkrp} \\
    \nonumber \nonumber \ \ \ &0 \le i < I, 0 \le j < J
\end{align}
 \vspace{-3mm}
 

\begin{table}[h]
\centering
\caption{Target problems for each target algorithm.}
\label{tab:problem_shapes}
\begin{tabular}{c|c|c|c|c|c}
    \hline
    \hline
    \textbf{CNN/MTTKRP} & \textbf{N/I} & \textbf{K/J} & \textbf{H,W/K} & \textbf{R,S} & \textbf{C/L} \\
    \hline
    \hline
    ResNet Conv\_3 & 16 & 128 & 28 &  3  &128\\ 
    ResNet Conv\_4 & 16 & 256 & 14 & 3 &256\\ 
    Inception Conv\_2 & 32 & 192 & 56  & 3 &192\\ 
    VGG Conv\_2 & 16 & 128 & 112 & 3  &64\\ 
    AlexNet Conv\_2 & 8 & 256 & 27  & 5  &96\\ 
    AlexNet Conv\_4 & 8 & 384 & 13  & 3  &384\\ 
    MTTKRP\_0 & 128 & 1024 & 4096 & - & 2048 \\
    MTTKRP\_1 & 2048 & 4096 & 1024 & - & 128 \\
    \hline
\end{tabular}
\end{table}

Table~\ref{tab:problem_shapes} shows the target problems we evaluated in Phase 2 
for each algorithm.
Specifically, we chose layers from popular networks such as ResNet~\cite{Resnet}, VGG~\cite{vgg}, AlexNet~\cite{Alexnet}, and Inception-V3~\cite{inception} and representative matrix shapes (tall and skinny~\cite{tall_skinny}) for MTTKRP.

\subsubsection{Hardware Accelerators}

We model the programmable hardware accelerator using Timeloop~\cite{timeloop}, which uses an analytical cost model to provide a high-fidelity cost estimation for hardware accelerators that implement affine loopnests.

The hardware accelerators we evaluate for both algorithms have the same memory hierarchy, namely
a two-level hierarchy with 512~KB of shared buffer and 64~KB of private buffer for each of 256 processing elements~(PEs). 
Buffers are banked and can be flexibly allocated to store any algorithm operand/partial result (tensors in the case of our target algorithms).
Each level of the memory hierarchy is coupled with control and address generation logic to support any loop order and tile size (similar to \cite{morph}).
The Network-on-Chip~(NoC) provides parallelism across the PEs along any combination of problem dimensions.

For each algorithm, we further specialize the datapath and control logic in PEs and the rest of the accelerator.
For CNN-Layer, PEs can consume 2 operands to produce 1 output per cycle, while for MTTKRP, PEs consume 3 operands to produce 1 output per cycle.
We assume the accelerator runs at 1~GHz and that the design objective is to minimize the energy-delay product~(EDP) to evaluate a problem.

\begin{figure*}[ht]
  \includegraphics[width=\textwidth]{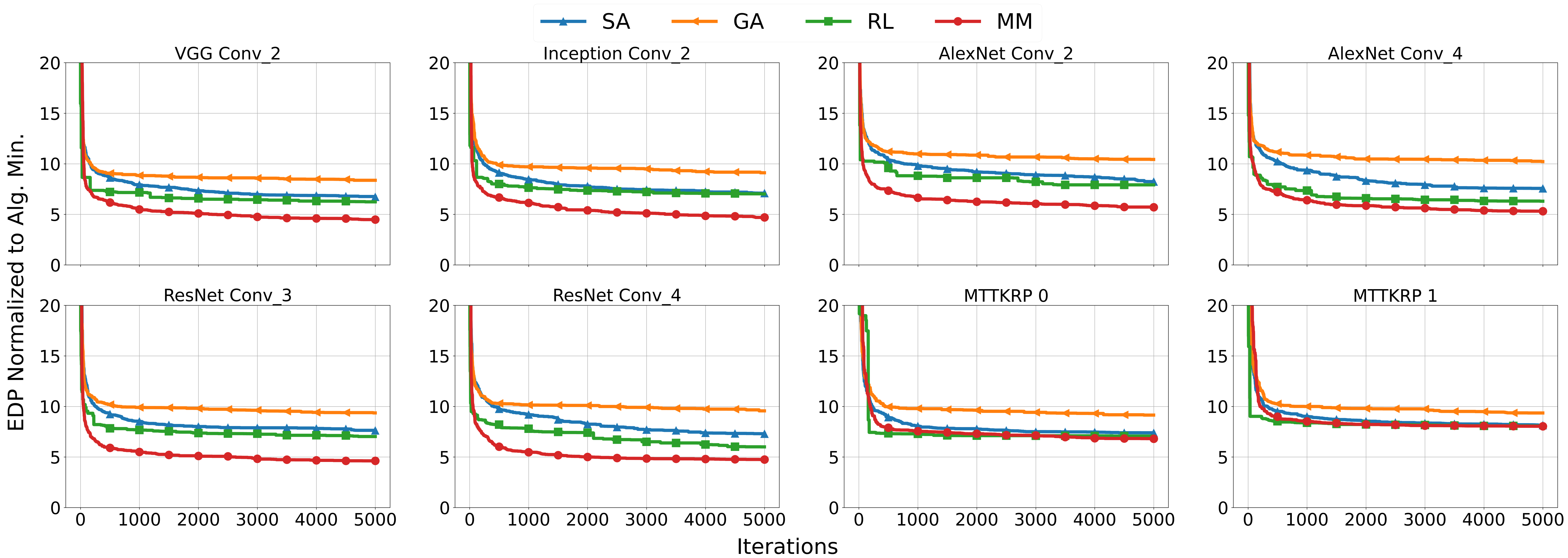}
\caption{Iso-iteration comparison of various search methods compared to Mind Mappings (MM).}
\label{fig:main_res_iters}
\end{figure*}

\subsubsection{Map Spaces}
\label{sec:eval:map_spaces}

Given the accelerator architecture $a$ and target problem $p$, each mapping $m \in M_{a,p}$~(Section~\ref{sec:method:input_rep}) 
is defined by the following programmable attributes for CNN-Layer/MTTKRP, typical in recent accelerators~\cite{eyeriss, scnn, ucnn, morph, flexiflow, maeri}.
\begin{enumerate}
    \item \textbf{Tiling:} The tile sizes for each dimension~(7/4) for each of the 3 levels in the memory hierarchy~(\emph{DRAM, L2, and L1}). (21/12 attributes for CNN-Layer and MTTKRP, respectively.)
    \item \textbf{Parallelism:} The degree of parallelism for each dimension across the PEs. (7/4 attributes.)
    \item \textbf{Loop Orders:} The ordering for each dimensions for each of the 3 memory hierarchies. (3/3 attributes.)
    \item \textbf{Buffer Allocation:} The allocation of banks for each tensor~(3/4) for 2 levels of on-chip memory hierarchy. (6/8 attributes.)
\end{enumerate}

These attributes induce a map space that is too large to exhaustively search.
For example, the map space size for the ResNet Conv\_4 layer~(CNN-Layer) is $\approx10^{25}$ valid mappings.

To characterize the search space, we sampled 1~M samples from each of the map spaces implied by Table~\ref{tab:problem_shapes} and computed the energy of each sample, which resulted in a $(mean, std)$ of $(44.2,231.4)$, $(48.0,51.2)$ for CNN-Layer/MTTKRP respectively, when energy 
was normalized to a theoretical lower-bound energy for the given problem.


\subsection{Search Methods and Comparison Metrics}
\label{subsec:eval:methods}
We compare Mind Mappings with following popular search methods used in prior work.
\begin{enumerate}
    \item \textbf{Algorithmic Minimum:} Refers to the theoretical lower-bound, possibly unachievable.
    \item \textbf{Simulated Annealing~(SA):} A popular black-box optimization method~\cite{simanneal}.
    \item \textbf{Genetic Algorithms~(GA):} Another popular black-box optimizer that uses evolutionary learning~\cite{genetic}.
    \item \textbf{Reinforcement Learning~(RL):} A popular unsupervised learning approach.
    \item \textbf{Mind Mappings~(MM):} This paper.
\end{enumerate}
We elaborate on how we implement each search method in Appendix~\ref{appendix:comparison}.

Two key metrics to judge a search method's effectiveness are the number of steps and amount of time needed to obtain an optimal solution.
Accordingly, we compare Mind Mappings against the above methods on two key metrics:
\begin{enumerate}
    \item \textbf{Iso-iteration Search Quality:} All approaches are run for fixed number of cost function evaluations. In case of Mind Mappings, the cost function is the trained surrogate~(Phase 1), whereas the other approaches query the actual cost function~(timeloop).
    \item \textbf{Iso-time Search Quality:} All approaches are run until a fixed wall-clock time.
\end{enumerate}

\begin{figure*}[ht]
  \includegraphics[width=\textwidth]{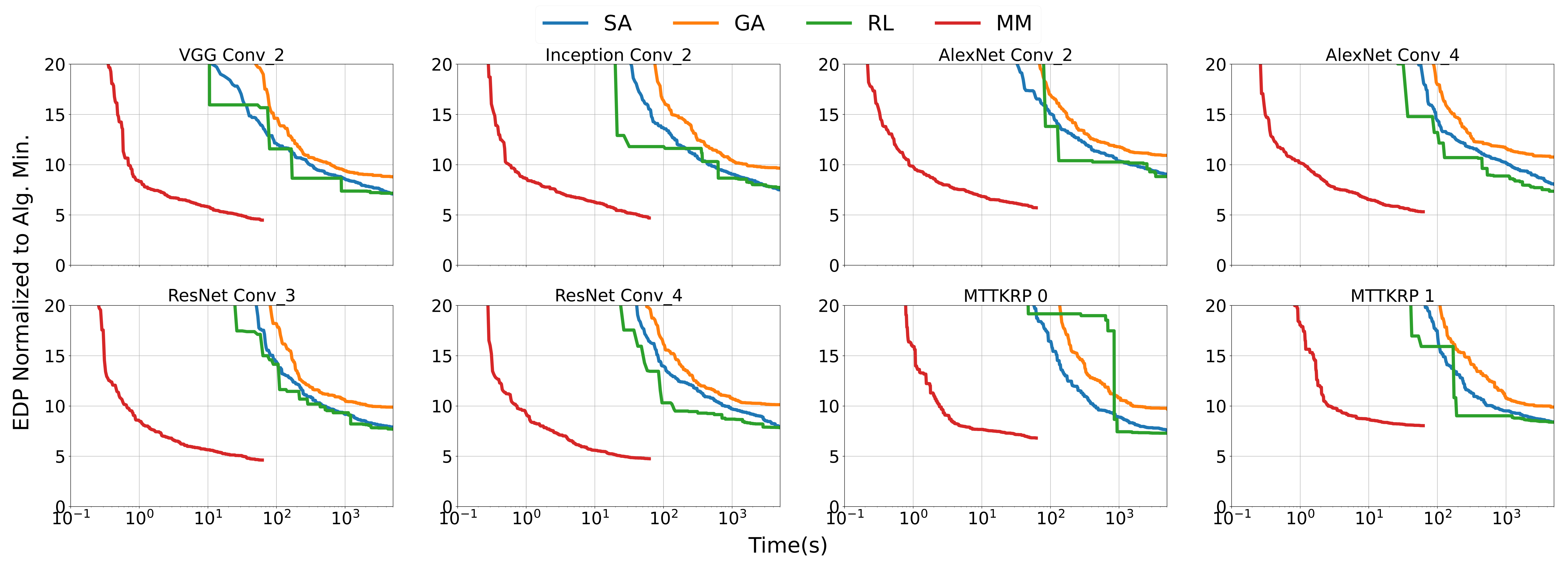}
\caption{Iso-time comparison of various search methods compared to Mind Mappings (MM).  Note the x-axis is log scale.}
\label{fig:main_res_time}
\end{figure*}

\subsection{Surrogate Model}
\label{subsec:eval:surrogate}

As discussed in Section~\ref{subsec:func_approx}, we implement the surrogate as a Multi-Layer Perceptron DNN.
We run Phase 1 once for each target algorithm.
That is, one surrogate is trained for all CNN-Layer results and a second is trained and used for all MTTKRP results.
This shows how the surrogate generalizes across problems for a given algorithm.
We elaborate on the details of the surrogate model and the training procedure in Section~\ref{subsec:eval:sensitivity}.




\subsection{Comparing Mind Mappings to Black-Box Optimization Approaches}

We plot iso-iteration and iso-time comparisons between Mind Mappings (MM) and the other search techniques~(Section~\ref{subsec:eval:methods}) for different problems~(Table~\ref{tab:problem_shapes}) in Figures~\ref{fig:main_res_iters} and \ref{fig:main_res_time}, respectively.
In both figures, the $y$-axis represents normalized EDP with respect to the algorithmic minimum for that problem~(Section~\ref{subsec:eval:methods}).
The $x$-axis represents iterations and time, respectively.
To isolate the effects of randomness, each method was run 100 times and the averaged results are plotted, i.e., for each iteration across the 100 runs, we average the EDP across the runs.

\subsubsection{Iso-iteration Comparison.}
\label{subsubsec:iso_iteration}
Figure~\ref{fig:main_res_iters} shows iso-iteration comparisons for all the search methods for every target problem represented in Table~\ref{tab:problem_shapes}.
Overall, MM outperforms SA, GA, and RL with mappings that have $1.40\times$, $1.76\times$, and $1.29\times$ lower EDP respectively, on average.
Further, solutions proposed by MM are only $5.3\times$ away from the (possibly unachievable) algorithmic minimum.
MM has a distinct advantage in that it performs a guided search using the approximate gradients derived from the surrogate model, where gradients by definition point at the steepest descent.

For CNN-layer based problems, 
MM converges to much better solutions compared to other approaches and does so within 1000 iterations.
The speed of convergence is the key characteristic that demonstrates the effectiveness of gradients and the guided nature of the search.
More importantly, MM performs well on every target problem~(layer shape for CNN-layer); 
indicating that the surrogate indeed generalizes and generates useful gradients across the family of problems associated with the target algorithm.

We note that for MTTKRP-based problems, MM converges to slightly better solutions compared to other approaches.
MTTKRP-based problems have much smaller map space sizes~($\approx10^{19}$ for MTTKRP\_0 vs. $\approx10^{25}$ for ResNet Conv\_4) and much lower variance in their EDP~(standard deviation of $51.2$ vs $231.4$ across a dataset of 1M samples; c.f. Section~\ref{sec:eval:map_spaces}), pointing to a possibly simpler mapping space search.
In such cases, black-box optimization approaches are competitive with MM in terms of iso-iteration search quality.
However, MM still provides good solutions much faster than other approaches in terms of iso-time search quality, as we will see in Section~\ref{subsubsec:iso_time}.

Other methods have known weaknesses, and sometimes map space search-specific weaknesses, that may be contributing to their lower search quality.
Specifically:
SA suffers from an inefficient traversal in the search space due to the unguided nature of search.
Not only is the performance of GA sensitive to the choice of initial population, but GA also heavily relies on an assumption that combining
two strong attributes in two mappings~(e.g., $m_d^i$ and $m_d^j$)
together will make an individual~($m'$) stronger. 
However, this is not necessarily
true in mapping space search.
For example, a good tiling for one loop order may not do well when combined with another loop order.
RL performs well compared to other black-box approaches, thanks to its superior heuristics using the actor-critic model based on deep deterministic policy gradient~(DDPG)~\cite{ddpg}.

\begin{figure*}[t]
  \centering
  \subfloat[Training and test loss.]{
	   \centering
        \includegraphics[width=0.30\textwidth]{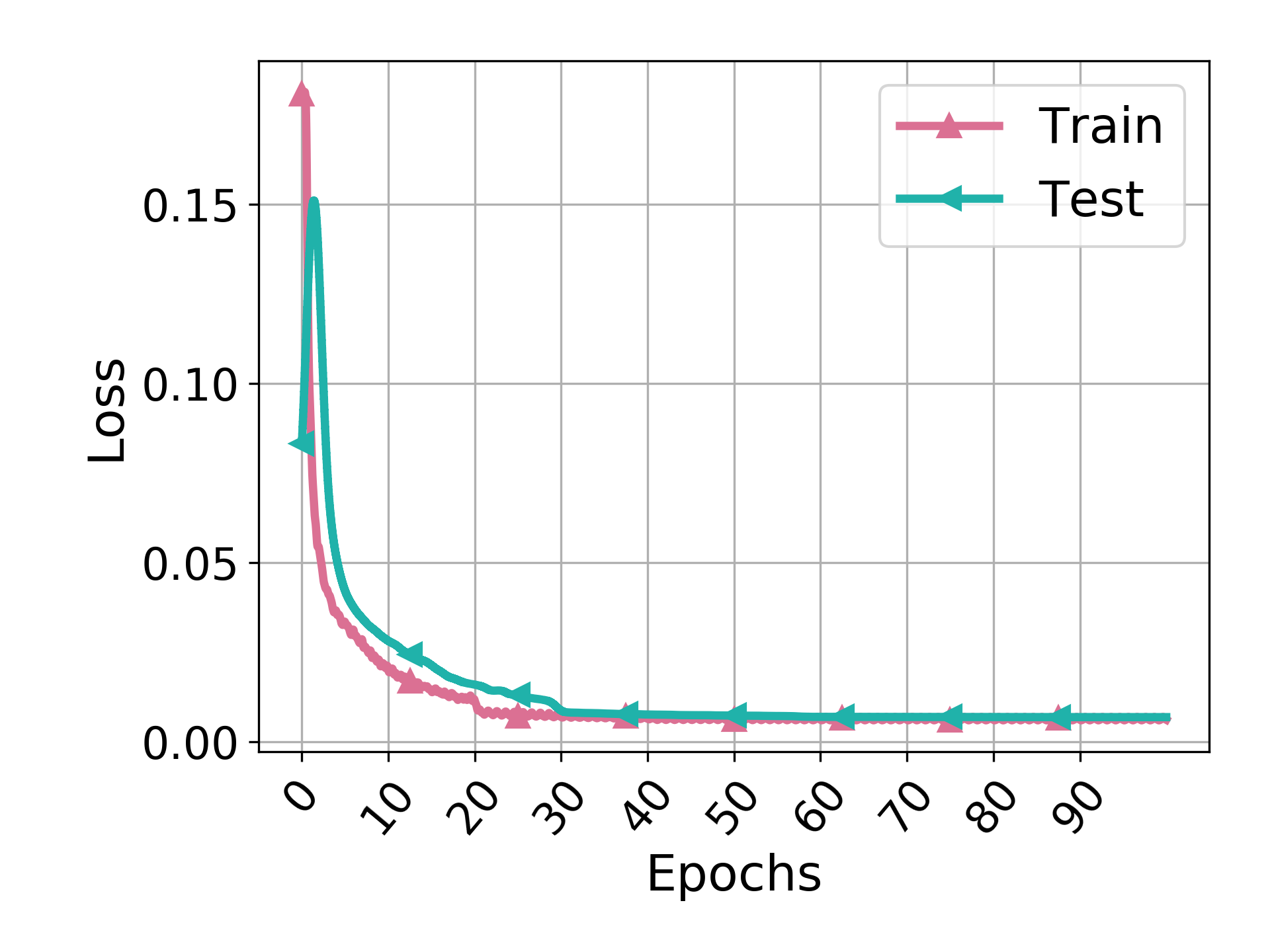}
        \label{fig:loss}
	}
 \hfill 	
  \subfloat[Choosing a loss function.]{
	   \centering
        \includegraphics[width=0.30\textwidth]{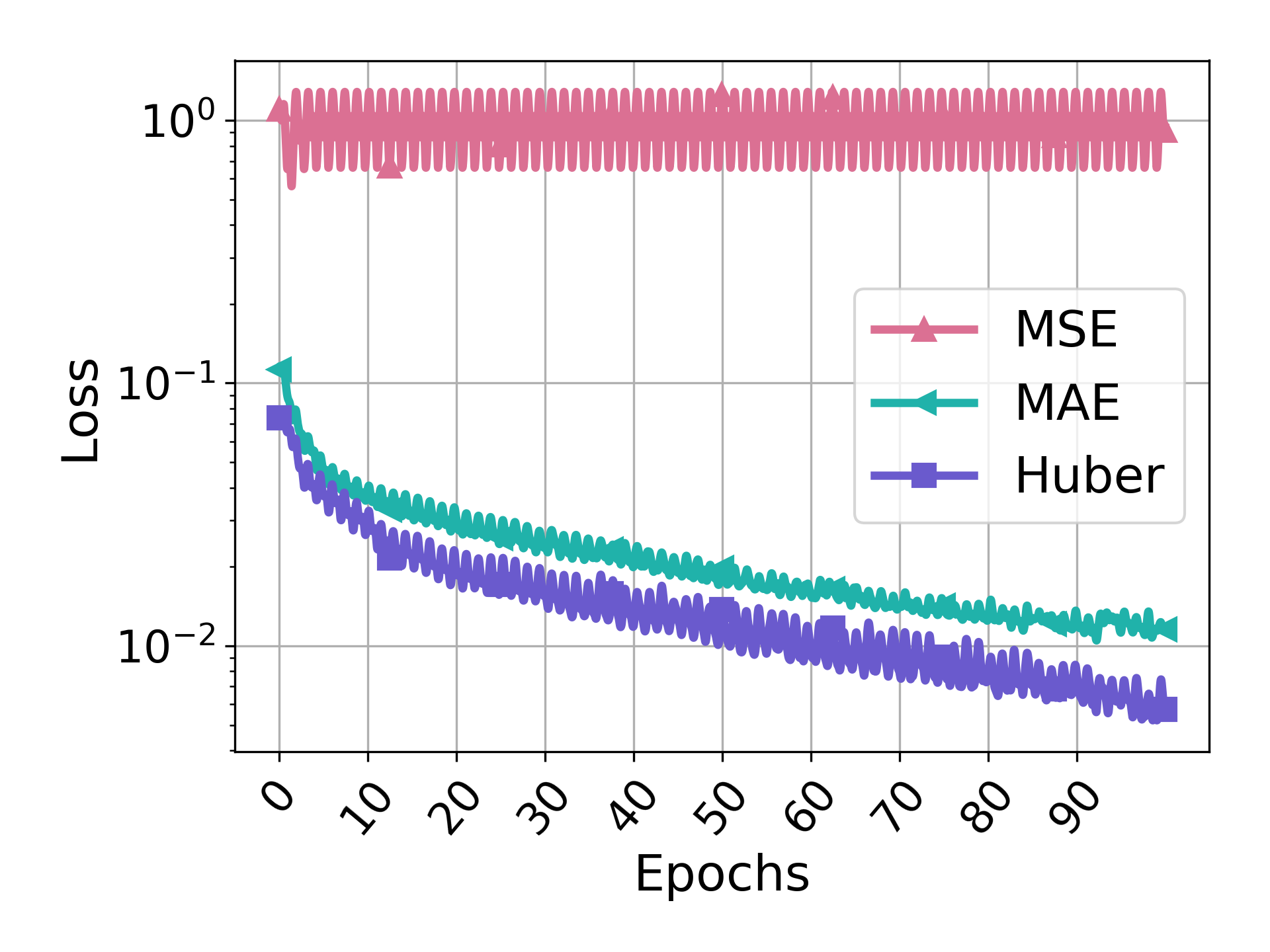}
        \label{fig:compare_loss}
    }
  \hfill
  \subfloat[Sensitivity to training set size.]{
	   \centering
        \includegraphics[width=0.30\textwidth]{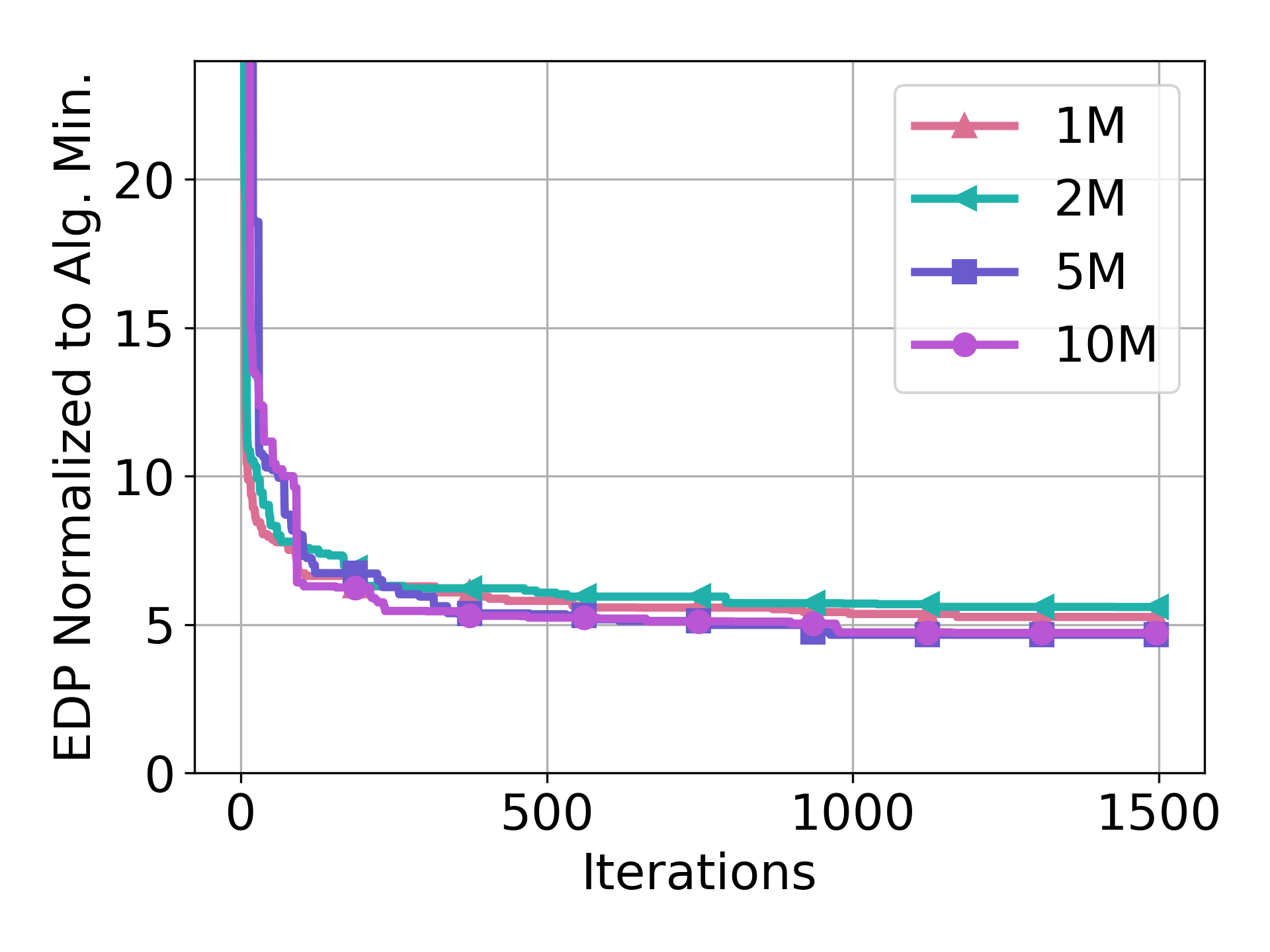}
        \label{fig:dataset_size}
	}
\caption{Experiments to determine the DNN topology and the loss function.}
\label{fig:sensitivity_train}
\end{figure*}

\subsubsection{Iso-time Comparison}
\label{subsubsec:iso_time}
Figure~\ref{fig:main_res_time} performs an iso-time study by plotting time as the $x$-axis~(note the log scale) when all search methods are run on an Intel Xeon E5-2637 v4 CPU.
Overall, MM outperforms SA, GA, and RL by $3.16\times$, $4.19\times$, and $2.90\times$ respectively on iso-time metric when run MM is run until convergence~(62.5 seconds).
This is possible since MM does not need to query the expensive cost function~(timeloop) every step, unlike other approaches.
Instead, MM uses the surrogate model to predict meta-statistics at every step ~(Section~\ref{sec:method:output_rep}), which in turn generates gradients to guide the next step.

Accordingly, we find that MM is 153.7$\times$, 286.8$\times$, and 425.5$\times$ faster per step than SA, GA, and RL respectively.
With recent hardware/software advances in Deep Learning, the gap between MM and other approaches can further widen by using state-of-the-art infrastructure instead of a CPU, as in this evaluation.
While RL outperforms SA/GA in iso-iteration search quality, the per step cost of RL is significantly higher than MM. 
To summarize: MM not only generates higher-quality mappings compared to other approaches, but generates those mappings faster. 


\textbf{Using Surrogate Models for Black-box Approaches.} We note that it is possible to improve traditional black-box methods in terms of time-per-step by using a surrogate, as explored in a several prior works~\cite{ithemal, tvm, release}~(refer to Section~\ref{subsec:surrogate_related}).
While such surrogates are not beneficial in finding better mappings (i.e., will not improve iso-iteration search quality), they enable more cost function queries per unit time, which improves iso-time search quality.

\subsubsection{Summary}

Overall, we note four key takeaways:

\begin{enumerate}
    \item \textbf{Generality:} Mind Mappings generalizes over different algorithms, architectures, and target problems, as demonstrated by the search performance.
    \item \textbf{Quality of Solution:} Mind Mappings finds better/as good mappings compared to other popular methods.
    \item \textbf{Optimality:} The Mind Mappings returns mappings that are within $5.3\times$ of the possibly unachievable lower bound, suggesting they are close to the achievable global optimum.
    \item \textbf{Time per Step:} Mind Mappings uses a surrogate model instead of the (expensive) accelerator cost function at every step, allowing it to perform more steps per unit time relative to other methods.
\end{enumerate}

\subsection{Surrogate Model} 
\label{subsec:eval:sensitivity}

We now provide details for the MLP used as the surrogate, and show sensitivity studies used to determine the MLP's training procedure and architecture.

\textbf{Input and Output Vectors.} 
The input mapping vector is 62/40 values in length for CNN-Layer/MTTKRP, respectively, which includes the representation of the problem id, tile sizes, parallelism, loop ordering, and buffer allocations.
We elaborate on the input vector representation for CNN-layer below.
\begin{enumerate}
    \item \textbf{Problem ID~($p_{id}$):} $P_0 = \mathbb{Z}^7$: a 7-tuple indicating the problem shape~($N,K,C,H,W,R,S$; see Table~\ref{tab:problem_shapes}).
    \item \textbf{Tile Sizes:} $P_1 = \mathbb{R}^{21}$: a 21-tuple representing what factor larger each tile dimension is relative to the corresponding dimension in the next level of the memory hierarchy (e.g., $R_c$ in the 1D-Conv tiled example in Code~\ref{code:tiled}).
    There are 21 factors, for the 7 dimensions in the 3-level memory hierarchy. 
    \item \textbf{Parallelism:} $P_2 = \mathbb{Z}^7$: a 7-tuple to indicate the degree of spatial parallelism for each dimension.
    Spatial parallelism is represented in terms of a factor for each dimension, similar to how tile sizes are represented (above).
    \item \textbf{Loop Order:} $P_3 = \mathbb{Z}^{21}$: a 21-tuple indicating the loop order, represented as a permutation of the 21 (7 dimensions times 3 levels of memory) loops.
    For example, in 1D-Conv $W\rightarrow R$ is represented as $[0,1]$, and $R\rightarrow W$ is $1,0$.
    \item \textbf{Buffer Allocation:} $P_4 = \mathbb{R}^6$: a 6-tuple indicating the percentage of banks in each of the 2 levels of on-chip memory allocated to each of the 3 tensors~($\inp$, $\out$, and $\filt$ in Equation~\ref{eq:conv2d}).
\end{enumerate}

The output cost vector has 12/15 neurons for CNN-Layer and MTTKRP, respectively. 
Each neuron represents the energy spent in accessing a specific level of the memory hierarchy~(3) for each input/output tensor~(3/4), overall energy, compute utilization, and overall cycles for execution.
Inputs and outputs are normalized over the dataset to have a mean of 0 and standard deviation of 1, as discussed in Section~\ref{subsec:func_approx}.

\textbf{DNN Topology and Training.}
We choose a 9-layer deep MLP with \lstinline{[64,256,1024,2048,2048,1024,256,64,12/15]} neurons in each layer for CNN-Layer/MTTKRP, respectively, as the surrogate model based on a grid search.
We train the MLP for 100 epochs with a learning rate of $10^{-2}$, which is decayed by a factor of 0.1 every 25 epochs, and a batch size of 128.
We use the Stochastic Gradient Descent (SGD) optimizer with a momentum value of 0.9.
Loss over the training duration is depicted in Figure~\ref{fig:loss}.
The MLP converges at around 60 epochs, and the test loss closely follows the train loss, indicating that we do not overfit.
 
\textbf{Dataset.}
We train the model with 10~M samples drawn with uniform random probability from the space of representative problems associated with the target algorithm~(Section~\ref{sec:method}).
``Representative problems'' means we sample from a range of typical values for each parameter making up the problem~(e.g., the $N,K,C,H,W,R,S$ dimensions for CNN-layer). 
For example, we randomly sample the value of $K$ for CNN-layer from the range \lstinline{[32,512]}, which should cover the typical range of $K$ in practice~\cite{Resnet,vgg,Alexnet}.
That Mind Mappings performs well given this methodology suggests that the surrogate is able to interpolate and predict costs for unseen combinations of problem parameters. 
Figure~\ref{fig:dataset_size} compares the search performance on surrogate models trained with 1~M, 2~M, 5~M, and 10~M samples.
While the datasets with more than 5~M samples lead to a well-trained model for this problem, surrogates used for the evaluation were trained with 10~M samples.
We note that, even when the dataset size is smaller than 5~M, search quality is not significantly hampered.


\textbf{Loss Function Choice.} We use the \emph{Huber} loss~\cite{huber} function as the loss criterion for training with the SGD optimizer.
Figure~\ref{fig:compare_loss} compares the performance of several popular loss functions used in regression such as the \emph{Mean Squared Error}~(MSE) and \emph{Mean Absolute Error}~(MAE).
Interestingly, MSE loss, a widely popular loss function used in regression problems, performs poorly in our setting.
We attribute this to the fact that MSE greatly punishes outliers, leading to a large variance in loss and instability in training.
By contrast, \emph{mean absolute error} punishes small variations, leading to sub-par performance.
Huber loss is similar to MSE when variations are small and is similar to MAE when the variations are larger, thus creating a good balance between the two loss functions.

\textbf{Model size.} When weights are represented as 32~bit floats, the surrogate model takes 35~MB of storage.
With recent advances in pruning and quantization~\cite{deepcompression, eie}, the model can be likely be compressed significantly.
Therefore, we believe that model size should not be a constraint in adopting Mind Mappings for real world applications.

%% file: ASPLOS CR/background.tex
\section{Related Work}
\label{sec:blackbox}

\begin{table*}[!t]
\centering
\caption{Related works in Mapping Space Search. Mind Mappings differentiates from other works by enabling a first-order optimization using Gradient Descent with a differentiable surrogate.
}
\begin{tabular}{c|c|c|c} 
 \hline
 \textbf{Work} & \textbf{Problem Domain} & \textbf{Cost Function} & \textbf{Search Heuristic}  \\ [0.5ex] 
 \hline\hline
 FlexTensor~\cite{flextensor} & Tensor Compilation & Actual Hardware & Reinforcement Learning \\
 Tiramisu~\cite{tiramisu} & DNN Compilation & Actual hardware & Beam Search \\
 Gamma~\cite{gamma} & DNN Mapping Space Search & Analytical & Genetic Algorithm \\
 TensorComprehensions~\cite{tensorComprehensions} & DNN Compilation & Actual hardware & Genetic Algorithms \\
   dMazeRunner~\cite{dmazerunner} & DNN Compilation & Analytical & Pruned Search \\
 Timeloop~\cite{timeloop} & Affine loop nests & Analytical & Pruned Search  \\
 TVM~\cite{tvm} & DNN Compilation & Gradient Boosted Trees & Simualted Annealing  \\
 RELEASE~\cite{release} & DNN Compilation & Gradient Boosted Trees\ & Reinforcement Learning \\ 
 Adams et. al~\cite{beamsearch_halide} & Halide~\cite{halide} Compilation & Multi-Layer Perceptrons & Beam Search \\

 \textbf{Mind Mappings (ours)} & \textbf{Domain Agnostic} & \textbf{Multi-Layer Perceptrons} & \textbf{Gradient-based  Search} \\ [1ex] 
 \hline
\end{tabular}
\label{table:related_work}
\end{table*}



We now describe prior work studying mapping space search as well as related search techniques applied to other problems.
We summarize the former in Table~\ref{table:related_work}.

\subsection{Mapping Space Search}
\label{subsec:surrogate_related}



To overcome the challenges in mapping search space, prior works use two main approaches:
~(i) reduce the time to evaluate the cost function, and
~(ii) avoid exhaustive search by adopting better heuristics.

\subsubsection{Faster Cost Estimation.}
\label{sec:related:fast_estimation}

For unguided mapping space search techniques that rely on exhaustive search or black-box methods, evaluating more mappings is the key to find higher-quality mappings.
However, a key challenge is that the cost to evaluate a mapping using the actual hardware or a representative simulator is non-trivial.
To get around this, prior works use several techniques, described below.


dMazeRunner~\cite{dmazerunner}, Timeloop~\cite{timeloop}, GAMMA~\cite{gamma}
use analytical models built by domain experts that are faster to evaluate than the actual hardware, and are sufficiently representative and flexible to support different mappings.
However, building such analytical models is difficult and requires strong domain expertise.
Some other works instead do away with domain expertise requirements by
leveraging machine learning to build an approximate cost function.
For example, AutoTVM~\cite{tvm} and RELEASE\cite{release} use gradient-boosted trees~\cite{xgboost}.
On the other hand, Adams et al.~\cite{beamsearch_halide} use Multi-layer Perceptrons (MLPs). 
We note that while we also use MLPs to build the surrogate,
we utilize the \emph{differentiability} of the surrogate to perform a guided search.


\subsubsection{Mapping Space Search with Heuristics.}

Orthogonal to techniques mentioned in Section~\ref{sec:related:fast_estimation} that speed up the cost evaluation, prior works also develop custom/learnt heuristics to improve the search itself, so as to avoid brute-force search.
dMazeRunner~\cite{dmazerunner}, Marvel~\cite{marvel} and Timeloop~\cite{timeloop} prune the search space to reduce the number of mappings that need to be evaluated using domain expert knowledge.
The key idea is that points in the search space can be eliminated without evaluation, e.g., tile sizes 
that do not fit in the on-chip buffer.
Again, this solution is difficult to scale 
since it requires extensive domain expertise to create rules to prune the search space.

Several prior works leverage black-box optimization methods to perform the search.
For example, AutoTVM uses parallel simulated annealing~\cite{simanneal}~(SA) to search through the map space. 
OpenTuner~\cite{opentuner} is a program auto-tuner that uses the AUC Bandit Meta technique to combine several methods such as differential evolution.
RELEASE~\cite{release} and FlexTensor~\cite{flextensor} both use Reinforcement Learning (RL) as the cost heuristic to guide the search.
Tiramisu~\cite{tiramisu} and Adams et al~\cite{beamsearch_halide} both employ beam search. 
Finally, TensorComprehensions~\cite{tensorComprehensions} and GAMMA~\cite{gamma} use Genetic Algorithms~\cite{ga_source}, which are a popular approach used in combinatorial optimization~\cite{ga_usage, ga_app_0,ga_app_1}.

For all of the above: 
by definition of being black box, heuristics can only guide the search based on previously visited samples, and therefore require a large number of samples to perform well.
As demonstrated in Section~\ref{sec:eval}, Mind Mappings outperforms SA, GA, and RL by utilizing powerful gradient-based optimization with the differentiable surrogate.



\subsection{Related Works in other Areas}
Beyond mapping space search, the combinatorial search represented in Equation~\ref{eq:schedule} is widely found in other areas such as neural architecture search~\cite{NAS_rl}, device placement~\cite{clite}, etc., and insights from related works in these areas can apply to the mapping space search problem.

\textbf{Surrogate Modeling.} Using surrogates for solving black-box optimization problems has been well explored~\cite{surrogate_modeling}.
To predict the performance of a program on a CPU, Ithermal~\cite{ithemal} and Difftune~\cite{difftune} use Recurrent Neural Networks, {\"I}pek et al.~\cite{surrogate_0} use Artificial Neural Nets, and Lee et al.~\cite{surrogate_1} use regression modeling.
Deep generative models are proposed as a surrogate in ~\cite{surrogate_2}, which are differentiable approximations.
Function approximation or surrogate modeling has been at the core of modern Reinforcement Learning methods to approximate the large state-space present in real-life problems.
Similarly, Mind Mappings uses a differentiable surrogate, while carefully tuning the supervised training methods to adapt to the mapping space search problem.


\textbf{Search Heuristics.}
Black-box approaches such as Simulated Annealing, Genetic Algorithms~\cite{ga_usage, ga_app_0,ga_app_1}, Bayesian Optimization~\cite{bayesian_dnn_0, clite, brandon}, etc., have been widely used in different applications.
Recent advances in Reinforcement Learning~(RL) have influenced several works~\cite{NAS_rl, haq} to adopt the same, with promising results.
Several works have explored gradient-based methods~\cite{fbnet, darts,surrogate_2, difftune}, in spite of the cost function being non-differentiable. For example, FBNet~\cite{fbnet} uses Gumbel-Softmax~\cite{gumbel} to make the discrete choices in their problem differentiable, thereby obtaining gradients.

While gradient-based optimization applied to combinatorial optimization problems with black-box cost functions is not new~\cite{fbnet, darts,surrogate_2, difftune, gradient_technique_0, gradient_technique_1, gradient_technique_2, gradient_technique_3, gradient_technique_4}, adapting this to the mapping space search problem---the focus of this paper---is new and faces non-trivial challenges (Section~\ref{sec:method}).

%% file: ASPLOS CR/conclusion.tex
\section{Conclusion}

This paper proposed Mind Mappings, an efficient method for performing algorithm-accelerator mapping space search.
The key idea is to approximate the non-differentiable accelerator cost function with a differentiable surrogate, and to use that surrogate to perform a powerful Gradient Descent-based search.

While Mind Mappings significantly closes the gap to finding optimal mappings quickly, there is still gap left to close.
In particular, Section~\ref{sec:method} details several areas where the method can be further optimized, ranging from improved sampling methods for training the surrogate to more efficient encodings of accelerator programmable attributes.
Long term and with these refinements, we hope that the methods in this paper advance mapping space search to a level closer to its more mature cousin, compilation for general-purpose devices.

%% file: ASPLOS CR/appendix.tex
\appendix

\begin{appendices}

\section{Evaluation: Points of Comparison}
\label{appendix:comparison}

We now provide implementation details for each search method used in Section~\ref{sec:eval}.

\textbf{Algorithmic Minimum.} Our baseline represents the theoretical lower-bound EDP for the given accelerator, algorithm and problem. 
We construct this oracle EDP by taking the product of the minimum energy and minimum execution cycles.
The minimum energy is achieved when each input data is read only once and each output data is written only once at each level in the memory hierarchy.
The minimum execution cycles are achieved when PEs maintain 100\% utilization, i.e., when cycles equals $required\_flops/(flops\_per\_pe * num\_pes)$.

Note that in practice, one usually trades-off energy for cycles and cannot achieve the best of both worlds.
Thus, the above algorithmic minimum is likely unachievable.
We do not calculate the achievable lower-bound EDP, as this requires an intractable exhaustive search.



\textbf{Simulated Annealing~(SA).}
We implement SA in Python using a popular library $simanneal$~\cite{simannealcode}.
For each problem evaluation, we let the library perform auto-tuning to get the best hyper-parameters for SA such as the temperature and annealing factor.
We interface the library with the Mind Mappings tuner to perform mapping space search.

\textbf{Genetic Algorithm~(GA).}
We implement GA in Python using DEAP~\cite{genetic:deap}, a popular GA library.
Based on the extensive literature on parameter tuning for GA~\cite{ga_tune_0,ga_tune_1,ga_tune_2,ga_tune_3,ga_tune_4} and a grid search, we set an initial population size of 100 and crossover/mutation probabilities of 0.75/0.05, respectively.
Each individual is a mapping ranked based on fitness, which represents the optimization objective, EDP.
Every iteration, we perform a cross-over and mutation over the population.
A cross-over results in swapping attributes of one individual with the other while a mutation is implemented as a .05 probability of a random update for each of the mapping's attributes.
At the end of each generation, individuals are chosen based on their fitness for the next generation.

\textbf{Reinforcement Learning~(RL).}
We implement RL in PyTorch~\cite{pytorch}, based on the Deep Deterministic Policy Gradient (DDPG)~\cite{ddpg} implementation from HAQ~\cite{haq}. In the RL setting, the mapping problem is modeled as a Markov Decision Process (MDP)~\cite{mdp}, 
where each mapping is a ~\emph{state} in the MDP, an \emph{action} results in a move to a target state and the cost of the mapping is the \emph{reward}.
In each episode, the RL agent starts from a random initial state, takes an action to move to a target state  and updates its \emph{policy} based on the reward.
In this process, the agent learns the optimal action to take given a state in the space. 
The learning process uses the actor-critic method~\cite{actor_critic}, which is a widely-used policy gradient algorithm. 
The actor and critic functions are approximated with two fully-connected DNNs with 300 neurons respectively. 


\textbf{Mind Mappings~(MM).}
We implement Mind Mappings (Section~\ref{sec:method}) using a trained surrogate model~(elaborated in Section~\ref{sec:design:mapping_search}) as described in Section~\ref{subsec:func_approx}.
We inject randomness at an interval of every 10 iterations 
to avoid local minimas, as described in Section~\ref{subsec:local_minima}.
We use simulated annealing with a temperature of 50 initially to decide the acceptance of random injections, which is annealed every 50 injections by a factor of 0.75.
We use a learning rate of 1, and we do not decay the learning rate throughout the procedure.
We choose the learning rates and injection interval via a grid search.

\section{Mind Mappings API}
\label{subsec:api}

The Mind Mappings API exposes an optimization framework for mapping space search that can be used in compilers and frameworks targeting a specialized hardware accelerator, such as TVM~\cite{tvm}, PyTorch~\cite{pytorch}, TensorFlow~\cite{tensorflow}, etc.
A surrogate model is trained offline for the target algorithm-accelerator pair to approximate mapping cost, using techniques described in Section~\ref{subsec:func_approx}.
Then, during the compilation, the Mind Mappings API takes the trained surrogate model and the target problem $p$ as input and returns a low-cost (ideally optimal) mapping $m_{opt}$ that minimizes the problem's execution cost on the given accelerator.

The Mind Mappings API requires the following routines:
(1) $\mathsf{getMapping}$: gives a random valid mapping, 
(2) $\mathsf{isMember}$: checks if a mapping is valid, and 
(3) $\mathsf{getProjection}$: returns a projection from an invalid mapping to the nearest valid mapping.
We have open sourced the Mind Mappings framework here: \url{https://github.com/kartik-hegde/mindMappings}.

\end{appendices}